\documentclass{article}
\usepackage{ijcai26}

\usepackage{times}
\usepackage{soul}
\usepackage{url}
\usepackage[hidelinks]{hyperref}
\usepackage[utf8]{inputenc}
\usepackage[small]{caption}
\usepackage{graphicx}
\usepackage{amsmath}
\usepackage{amsthm}
\usepackage{booktabs}
\usepackage{algorithm}
\usepackage{algorithmic}
\usepackage[switch]{lineno}
\usepackage{multirow}
\usepackage{tabularx}
\usepackage{amssymb}

\title{SDFLoRA: Selective Decoupled Federated LoRA for Privacy-preserving Fine-tuning with Heterogeneous Clients}

\author{
Zhikang Shen$^1$
\and
Jianrong Lu$^1$
\and
Haiyuan Wan$^2$
\And
Jianhai Chen$^{1,}$\footnote{Corresponding author.}
\affiliations
$^1$Zhejiang University\\
$^2$Tsinghua University
\emails
\{hydrolysis02, jrong.alvin\}@gmail.com,
wanhy24@mails.tsinghua.edu.cn,
chenjh919@zju.edu.cn
}

\begin{document}

\maketitle

\begin{abstract}
    Federated learning (FL) has emerged as a promising paradigm for adapting large language models (LLMs) to distributed data. To mitigate the high communication and memory overhead, parameter-efficient techniques such as Low Rank Adaptation (LoRA) are widely adopted. However, practical deployments often exhibit \emph{rank and data heterogeneity}, making direct aggregation of LoRA updates biased and unstable. Existing approaches enforce a unified rank or align heterogeneous updates into a single shared subspace, which undermines personalization and accuracy. Moreover, under differential privacy (DP), adding noise to such mixed updates could perturb client-specific directions that should remain local, resulting in utility loss. To address these issues, we propose \textbf{S}elective \textbf{D}ecoupled \textbf{F}ederated \textbf{LoRA} (\textbf{\textit{SDFLoRA}}), a structure-aware LoRA framework that decouples each client update into a shared component for updating, and a private component that preserves client-specific semantics. Subspace alignment and aggregation are applied selectively to the shared module, while private modules remain local. We further design low rank re-compression to maintain a fixed rank budget for the shared module. This structure supports privacy aware optimization by injecting DP noise exclusively into the shared module. Experiments on multiple benchmarks demonstrate that SDFLoRA outperforms federated LoRA baselines and exhibits strong robustness under privacy constraints.
\end{abstract}

\section{Introduction}

Parameter-efficient federated fine-tuning (PEFT) has emerged as a critical paradigm for adapting Large Language Models (LLMs) to distributed, privacy-sensitive datasets.
By optimizing only lightweight adapters, clients avoid full-model communication and memory overhead, enabling deployments in domains such as healthcare, finance, and on-device intelligence.
Among PEFT approaches, low-rank adaptation (LoRA)~\cite{hu2022lora} is the dominant choice in federated LLM adaptation and is widely adopted in recent FL-LLM systems~\cite{wang2024flora,yang2025federated,long2024dual}.

Despite these advances, practical Federated Learning (FL) deployments could exhibit strong heterogeneity, as clients originate from different domains, tasks, and user populations. LoRA updates naturally contain both transferable directions shared across clients and client-specific directions that encode local semantics and personalization.
However, existing federated LoRA pipelines typically enforce shared alignment and aggregation over all update directions, which suppresses client-specific adaptations and limits personalization.
This issue is further exacerbated by rank heterogeneity, where clients adopt different low rank configurations due to diverse system constraints, making naïve aggregation even more biased and unstable.

Existing solutions for federated LoRA under heterogeneous ranks can be grouped into two paradigms.
The first paradigm enforces a unified rank across all clients, which reduces flexibility and can waste resources on constrained devices~\cite{cho2024heterogeneous}.
The second paradigm aligns heterogeneous LoRA updates into a shared subspace to enable aggregation. However, it tends to align all directions indiscriminately, including client-specific components that should remain private and unshared, thereby harming personalization~\cite{zhao2023aggregation}.
Meanwhile, differential privacy (DP) is widely adopted in FL to protect private client information.
When such structurally entangled updates are globally aligned, adding noise to the entire update could perturb client-specific components that are not meant to be shared, leading to utility loss without delivering a meaningful privacy guarantee~\cite{wei2023personalized}.

Motivated by these observations, we propose \textbf{S}elective \textbf{D}ecoupled \textbf{F}ederated \textbf{LoRA} (\textbf{\textit{SDFLoRA}}), which decomposes client adapter into a shared module that captures transferable knowledge and a private module that preserves client-specific adaptations.
Only the shared module participates in selective subspace alignment and aggregation across clients, while private modules remain private and unaggregated. To control rank growth induced by stacking-based aggregation, we further introduce low rank re-compression to maintain a fixed rank budget for the shared module. This structure naturally supports privacy aware optimization by applying DP mechanisms exclusively to the shared module updates, avoiding unnecessary perturbations on local adaptations.

Our contributions are summarized as follows:
\begin{itemize}
    \item We propose SDFLoRA, a structure-aware federated LoRA framework that decouples each client adapter into a shared module and a private module, and performs selective subspace alignment and aggregation exclusively on the shared module.
    The shared module is further constrained by a fixed rank budget via low rank re-compression, enabling stable and efficient aggregation.
    \item We introduce a DP compatible optimization enabled by the decoupling, where differential privacy noise is injected only into the aggregated shareable update. This design avoids unnecessary perturbations on client-specific directions and leads to improved robustness under privacy constraints.
    \item Extensive experiments on multiple benchmarks with heterogeneous clients demonstrate that SDFLoRA outperforms representative federated LoRA baselines, particularly in achieving a superior balance between privacy protection and model performance.
\end{itemize}

\section{Related Work}

\subsection{Federated Optimization under Client Heterogeneity}

FL enables collaborative model training without centralizing data by alternating between local optimization and server-side aggregation.
Federated Averaging (FedAvg)~\cite{mcmahan2017communication} is the canonical approach, but its simple weighted averaging mechanism often leads to unstable convergence under non-IID data distributions, partial client participation, and multiple local update steps, a phenomenon commonly referred to as client drift.

To mitigate heterogeneity-induced optimization challenges, a lot of work has proposed algorithmic modifications from different perspectives.
FedProx~\cite{li2020federated} introduces a proximal regularization term to constrain local updates from deviating excessively from the shared model.
FedDyn~\cite{durmus2021federated} and FedDC~\cite{gao2022feddc} further address client drift by incorporating dynamic correction terms that explicitly decouple local objectives from the shared optimization trajectory.
Another line of work focuses on correcting biases caused by heterogeneous local computation.
FedNova~\cite{wang2020tackling} normalizes client updates according to their effective number of local steps, while SCAFFOLD~\cite{karimireddy2020scaffold} reduces update variance through control variates shared between the server and clients.
From an optimization-centric viewpoint, MIME and MIME-Lite~\cite{karimireddy2020mime} reinterpret federated learning as biased stochastic optimization and improve convergence by aligning federated updates with centralized SGD via momentum and gradient correction.

Despite their demonstrated effectiveness, these methods are predominantly designed for full-parameter federated optimization and implicitly assume homogeneous model structures across clients.
When parameter-efficient fine-tuning is adopted, additional forms of heterogeneity emerge.
In particular, low-rank adapters introduce \emph{structural heterogeneity}, where clients may employ different adapter ranks or decompositions due to resource constraints or task diversity, which cannot be directly handled by existing optimization-centric approaches.

\subsection{Parameter-Efficient Fine-Tuning and Federated LoRA}

PEFT has become a standard approach for adapting large models under resource constraints.
LoRA~\cite{hu2022lora} freezes the backbone and learns low-rank updates for selected layers, greatly reducing trainable parameters and communication costs, which makes it especially attractive in FL.

Several studies extend LoRA to federated settings.
FedIT~\cite{zhang2024towards} performs FedAvg-style aggregation over LoRA parameters, enabling efficient collaboration but implicitly assumes homogeneous adapter configurations and can be sensitive to rank mismatch.
To mitigate rank heterogeneity, recent work has explored \emph{stacking-based aggregation} to align heterogeneous low-rank updates into a shared subspace.
FLoRA~\cite{wang2024flora} is a representative approach that constructs an expanded low-rank subspace via stacking, providing a principled way to aggregate clients with different ranks without padding or truncation.
Other variants improve initialization, aggregation bias, or fairness properties for federated LoRA training~\cite{bian2025lora,guoselective}.

These methods demonstrate that stacking is an effective tool for handling \emph{rank heterogeneity}.
However, they typically treat each client adapter as a monolithic update and implicitly assume that \emph{all} low-rank directions should be globally aligned and aggregated.
In federated LLM fine-tuning, client updates may contain a mixture of transferable knowledge and locally specific adaptations; indiscriminate alignment can over-regularize personalization and become fragile when additional constraints (e.g., privacy noise) are introduced.

\subsection{Personalization and Privacy in Federated Learning}
Personalized FL aims to account for client-specific objectives and data distributions by learning both shared and private components.
Representative methods include Ditto~\cite{gao2024feddtg}, which couples a shared model with personalized local models via regularization, and FedRep~\cite{collins2021exploiting}, which separates shared representations from client-specific heads.
These approaches highlight the importance of explicitly distinguishing between shared and locally specialized parameters, but they are not designed around the structural properties of low-rank adapters in LLM fine-tuning.

Privacy preservation is another fundamental concern in FL.
Differential privacy (DP)~\cite{dwork2006differential} provides formal guarantees, and DP-SGD~\cite{abadi2016deep} has become a standard mechanism to protect training updates by gradient clipping and noise injection.
In federated settings, differential privacy is typically enforced through
client-side or server-side perturbations, such as DP-SGD-based noise injection on local updates or aggregated models~\cite{kairouz2021advances}.
However, applying DP noise uniformly to all parameters can significantly degrade utility, particularly for large models and heterogeneous clients, and adversarial or poisoned updates further exacerbate this instability in federated aggregation~\cite{10417061}.
Recent work suggests that selective or structured privacy mechanisms may yield better utility--privacy trade-offs, yet existing federated LoRA approaches do not explicitly leverage adapter structure to support such selective protection.

\subsection{Positioning of Our Work}
Compared to prior federated LoRA methods, our work focuses on the semantic roles of low-rank updates and the resulting need for structural awareness in aggregation.
Rather than proposing another stacking mechanism, we introduce SDFLoRA that separates globally alignable updates from locally personalized ones.
Stacking-based aggregation is then applied \emph{selectively} to the shared module, while private modules remain private and unaggregated.
This perspective addresses not only how to aggregate heterogeneous ranks, but also which updates should be aggregated under personalization and privacy constraints, enabling robust federated LLM fine-tuning with improved stability under DP noise.

\section{Method}
\label{sec:method}

\subsection{Preliminaries: Federated LoRA Fine-tuning}
\label{sec:prelim}
We consider federated fine-tuning with $K$ clients. In round $t$, a subset $S_t$ of clients participates, each optimizing a local objective $F_k(\cdot)$ on its private data. Let $W$ denote the frozen backbone parameters of an LLM. LoRA~\cite{hu2022lora} augments a target linear layer $W \in \mathbb{R}^{d_{\text{out}}\times d_{\text{in}}}$ with a low-rank update
\begin{equation}
\Delta W_k = B_k A_k,\quad 
A_k \in \mathbb{R}^{r_k \times d_{\text{in}}},\;
B_k \in \mathbb{R}^{d_{\text{out}} \times r_k}
\label{eq:lora_update}
\end{equation}
where $r_k$ is the adapter rank (potentially different across clients). During local training, client $k$ updates $\{A_k,B_k\}$ (and keeps $W$ fixed). The server aggregates client updates and broadcasts the aggregated adapter for the next round.

\paragraph{Challenge: rank heterogeneity and semantic heterogeneity.}
When ranks differ across clients, directly averaging low-rank factors is ill-defined without padding or truncation, and may introduce biased updates. Stacking-based aggregation alleviates dimensional mismatch by aligning heterogeneous updates into an expanded subspace, but implicitly assumes that \emph{all} updates should be globally aligned and aggregated. In federated LLM fine-tuning, client updates often contain a mixture of globally transferable knowledge and client-specific adaptations; forcing shared alignment on all updates can harm personalization and becomes fragile under differential privacy noise. Our goal is therefore to design an aggregation scheme that is robust to rank heterogeneity while respecting the semantic roles of low-rank updates.

\subsection{Dual-Model LoRA Decomposition}
\label{sec:dual_module}
We propose to decompose each client adapter into two structurally separated low-rank modules:
\begin{equation}
\Delta W_k = \Delta W_k^{(g)} + \Delta W_k^{(l)}
          = B_k^{(g)}A_k^{(g)} + B_k^{(l)}A_k^{(l)}
\label{eq:dual_module}
\end{equation}
where $\Delta W_k^{(g)}$ is a \textbf{shared module} intended to capture transferable directions shared across clients, and
$\Delta W_k^{(l)}$ is a \textbf{private module} for client-specific adaptations.

This decomposition reflects an important observation in federated LLM fine-tuning: client-side low-rank updates are \emph{semantically heterogeneous}.
Some update directions correspond to transferable knowledge that should be aligned and shared across clients, while others encode client-specific or domain-specific adaptations whose forced alignment can be harmful.
By explicitly separating $\Delta W_k^{(g)}$ and $\Delta W_k^{(l)}$,
we impose a structural constraint on the subspaces that are allowed to participate in cross-client aggregation.

\begin{figure*}[t]
    \centering
    \includegraphics[width=\textwidth]{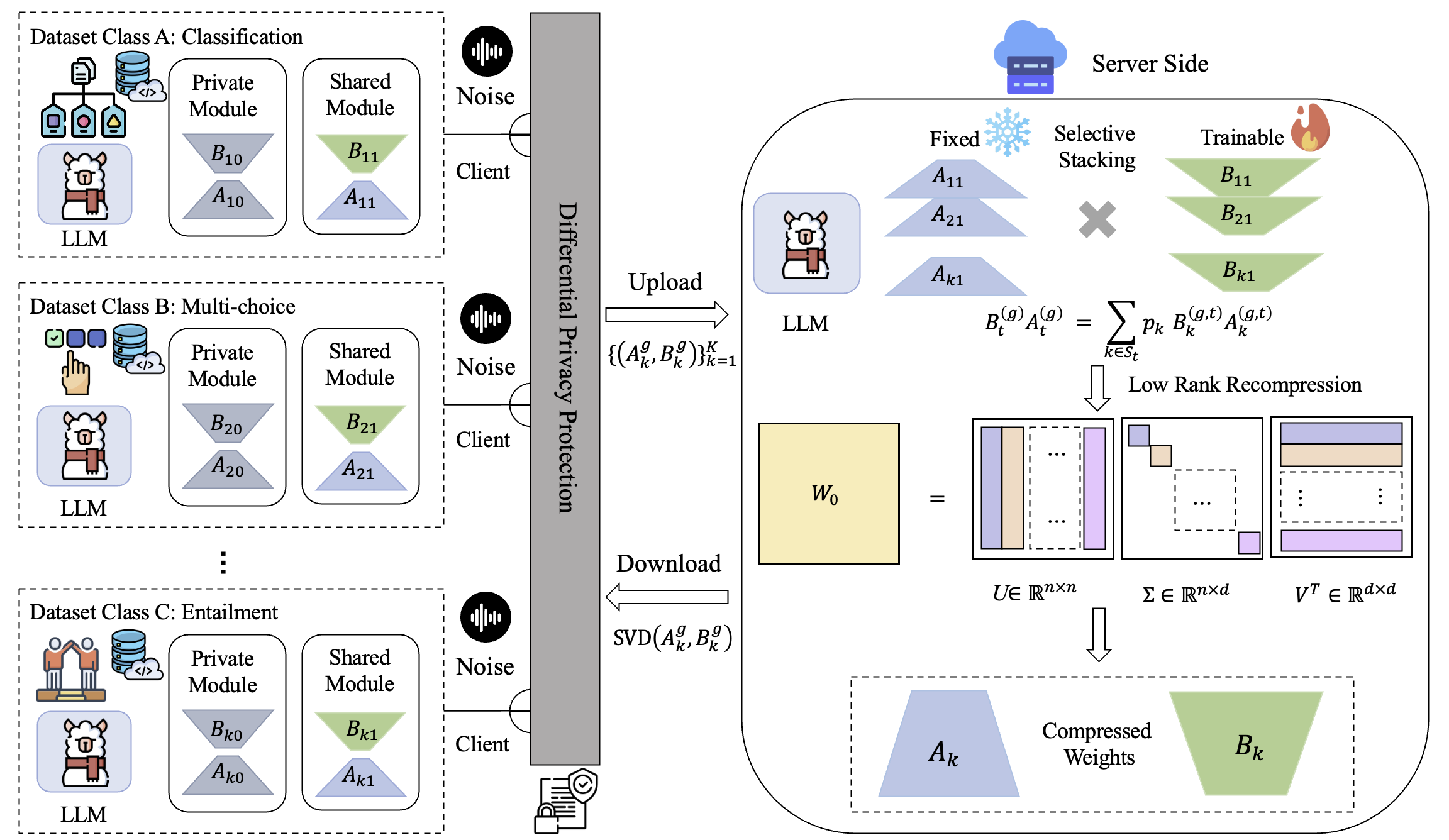}
    \caption{Framework overview. Each client keeps a private module and trains a shared module on a frozen LLM. Only the shared module is communicated and aggregated via selective stacking; Differential Privacy noise is applied to the shared update.}
    \label{fig:framework}
\end{figure*}

As illustrated in Figure~\ref{fig:framework}, we decouple transferable (public) updates from client-specific (private) adaptations, enabling selective aggregation and privacy-preserving training.

\paragraph{Federated roles.}
The two modules play different roles in the federated process:
(i) the shared module participates in cross-client aggregation and is broadcast each round;
(ii) the private module is \emph{never aggregated} and remains private to the client.
This structural separation is the key design choice that allows us to restrict subspace alignment to semantically appropriate updates, rather than treating each LoRA adapter as a monolithic update.

\paragraph{Optimization objective.}
In each round $t$, client $k \in S_t$ receives the current shared module $\theta^{(g,t)}$ (the collection of all $\{A^{(g)},B^{(g)}\}$ across adapted layers) and keeps its private module $\theta_k^{(l,t)}$.
The client performs $E$ steps of local optimization to minimize:
\begin{equation}
\min_{\theta^{(g)},\,\theta_k^{(l)}}\;
F_k\!\big(W,\theta^{(g)},\theta_k^{(l)}\big)
\label{eq:local_obj}
\end{equation}
and uploads only the updated shared module $\theta_k^{(g,t)}$ (or its update) to the server. The private module is kept on-device and continues across rounds.

\subsection{Selective Stacking Aggregation on the Shared Module}
\label{sec:selective_stacking}
We propose \emph{selective stacking}, a stacking-based aggregation strategy
that is applied exclusively to the shared module while explicitly excluding
client-private updates from cross-client alignment.
Let $\theta_k^{(g,t)}$ denote the shared-module parameters trained by client $k$ at round $t$ for a given layer, i.e., $\theta_k^{(g,t)}=\{A_k^{(g,t)},B_k^{(g,t)}\}$ with rank $r_k^{(g)}$.
To aggregate heterogeneous ranks without padding or truncation, we perform \textbf{stacking-based} aggregation only on the shared module.

\paragraph{Stacking construction.}
For a target layer, we form a stacked representation:
\begin{equation}
A^{(g)}_{t} = \mathrm{Concat}\!\left(p_k\,A_k^{(g,t)}\right)_{k\in S_t}
\label{eq:stacking}
\end{equation}
\begin{equation}
B^{(g)}_{t} = \mathrm{Concat}\!\left(p_k\,B_k^{(g,t)}\right)_{k\in S_t}
\label{eq:stacking}
\end{equation}
where $p_k$ is the aggregation weight (e.g., proportional to the local data size) and $\mathrm{Concat}(\cdot)$ concatenates along the rank dimension. This yields an aggregated update
\begin{equation}
\Delta W_t^{(g)} = B^{(g)}_{t}A^{(g)}_{t}
\label{eq:shared_update}
\end{equation}
which is well-defined under rank heterogeneity. Importantly, \emph{private modules are excluded from stacking and aggregation}, avoiding forced alignment of client-specific directions.

The stacking operation preserves the following equivalence:
\begin{equation}
B_t^{(g)} A_t^{(g)} = \sum_{k \in S_t} p_k B_k^{(g,t)} A_k^{(g,t)}
\end{equation}
while avoiding direct averaging of incompatible low-rank factors.
In contrast to padding-based alignment, stacking delays subspace merging
until after aggregation, preventing premature mixing of heterogeneous update directions.

\paragraph{Rank control by re-compression.}
A practical consequence of stacking is that the effective rank of the aggregated shared update grows with the number of participating clients and communication rounds.
Concretely, the concatenation in Eqs.~(\ref{eq:stacking})--(\ref{eq:shared_update}) expands the rank dimension, which increases both the transmission cost (larger adapter factors) and the memory footprint on the server and clients.

Moreover, under privacy-aware training, stochasticity (e.g., DP noise) and client heterogeneity tend to inject low-energy directions that accumulate across rounds, leading to unstable optimization and diminished generalization if left uncontrolled.
To keep the shared module within a fixed budget, we apply low-rank re-compression to $\Delta W_t^{(g)}$:
\begin{equation}
\Delta W_t^{(g)} \approx \tilde{B}_t^{(g)}\tilde{A}_t^{(g)},\quad
\mathrm{rank}(\tilde{B}_t^{(g)}\tilde{A}_t^{(g)})=r_{\max}.
\label{eq:recompress}
\end{equation}
In practice, we compute a truncated SVD (or an equivalent low-rank factorization)
$\Delta W_t^{(g)} = U\Sigma V^\top$ and retain only the top-$r_{\max}$ singular components,
yielding $\Delta W_t^{(g)} \approx U_{r}\Sigma_{r}V_{r}^\top$.
We then convert this truncated form back to LoRA-style factors by setting
$\tilde{B}_t^{(g)} = U_{r}\Sigma_{r}^{1/2}$ and $\tilde{A}_t^{(g)}=\Sigma_{r}^{1/2}V_{r}^\top$,
so that $\tilde{B}_t^{(g)}\tilde{A}_t^{(g)}$ matches the best rank-$r_{\max}$ approximation under the Frobenius norm.
This step acts as a spectral filter that preserves dominant, transferable directions while discarding tail components that are more likely to be noise-dominated, and it keeps the communicated shared module compact for the next round.
Finally, the server broadcasts $(\tilde{A}_t^{(g)}, \tilde{B}_t^{(g)})$ as the next-round shared module.

\subsection{Privacy Aware Optimization (DP Compatibility)}
\label{sec:dp}
Our decoupled design is naturally compatible with differential privacy (DP). The key idea is to apply DP mechanisms \emph{only} to the shared module, while keeping private modules noise-free.

\paragraph{Client-side DP-SGD on the shared module.}
During local training, client $k$ clips per-sample (or per-microbatch) gradients for the shared module and adds Gaussian noise:
\begin{equation}
g_k^{(g)} \leftarrow \mathrm{Clip}\!\left(g_k^{(g)}, C\right) + \mathcal{N}\!\left(0,\sigma^2 C^2 I\right)
\label{eq:dpsgd}
\end{equation}
where $C$ is the clipping norm and $\sigma$ is the noise multiplier. The perturbed gradients update $\theta_k^{(g)}$, which is then uploaded to the server. The private module $\theta_k^{(l)}$ is updated without noise, preserving client-specific expressiveness and improving optimization stability.

\paragraph{Why selective privacy helps.}
Injecting noise into the entire adapter amplifies variance and can destabilize learning, especially for heterogeneous clients. By restricting DP to the shared module, our framework protects transferable knowledge while avoiding unnecessary corruption of client-specific adaptations, yielding a better balance of utility and privacy.

\subsection{Algorithm Summary}
\label{sec:algo_summary}
Algorithm~\ref{alg:dual_selective} summarizes one communication round of our method. Each client receives the shared module, performs local training with decoupled adapters, uploads only the shared-module parameters, and the server performs selective stacking aggregation to produce the next shared module.

\begin{algorithm}[tb]
\caption{Selective Decoupled Federated LoRA (One Round)}
\label{alg:dual_selective}
\begin{algorithmic}[1]
\STATE \textbf{Input:} Frozen backbone $W$; shared module $\theta^{(g,t)}$; client set $S_t$;
weights $\{p_k\}$; rank budget $r_{\max}$; DP parameters $(C,\sigma)$
\STATE \textbf{Output:} Updated shared module $\theta^{(g,t+1)}$

\STATE Server broadcasts $\theta^{(g,t)}$ to clients $k \in S_t$.
\FOR{each client $k \in S_t$ \textbf{in parallel}}
    \STATE Initialize decoupled adapters with $\theta^{(g,t)}$ and local state $\theta_k^{(l,t)}$.
    \FOR{local step $e = 1,\dots,E$}
        \STATE Compute gradients on batch; update $\theta_k^{(g)}$
        (with DP-SGD if enabled) and $\theta_k^{(l)}$.
    \ENDFOR
    \STATE Upload $\theta_k^{(g,t)}$ to server; keep $\theta_k^{(l,t)}$ locally.
\ENDFOR
\STATE Server aggregates shared modules via selective stacking:
\STATE \hspace{1em} $\theta^{(g)} \leftarrow \mathrm{StackAgg}
(\{\theta_k^{(g,t)}\}_{k\in S_t}, \{p_k\})$.
\STATE Re-compress to rank budget $r_{\max}$ to obtain $\theta^{(g,t+1)}$;
otherwise set $\theta^{(g,t+1)} \leftarrow \theta^{(g)}$.
\end{algorithmic}
\end{algorithm}

\section{Experiments}
\label{sec:experiments}

\subsection{Experimental Setup and Baselines}
\label{sec:exp_setup}

\begin{table}[t]
\centering
\setlength{\tabcolsep}{5pt}
\renewcommand{\arraystretch}{1.1}
\begin{tabularx}{\columnwidth}{@{}lX@{}}
\toprule
Item & Setting \\
\midrule
Backbone        & LLaMA-7B \\
Adapter         & LoRA \\
Tasks           & GLUE (RTE, QQP, MNLI, SST-2), MMLU \\
Clients ($K$)   & 8 \\
Rounds ($T$)    & 30 \\
Rank (hetero.)  & $\{r_k\}=[4,4,8,8,8,8,16,16]$ \\
\midrule
Training        & Optimizer: Adam; learning rate $2\times10^{-4}$ \\
                & Local steps per round: 10; batch size: 128 \\
\midrule
Hardware        & Intel Xeon Gold 5318Y; 128GB RAM \\
                & 2$\times$ RTX A6000 \\
Software        & Ubuntu 22.04.5; Python 3.10; CUDA 12.4 \\
\bottomrule
\end{tabularx}
\caption{Experimental configuration and system environment.}
\label{tab:exp_config}
\end{table}

\begin{table*}[t]
\centering
\small
\setlength{\tabcolsep}{4.2pt}
\renewcommand{\arraystretch}{1.08}
\begin{tabular}{@{}l c c c c c c c c c c c c c@{}}
\toprule
& \multicolumn{4}{c}{\textbf{Hetero}}
& \multicolumn{9}{c}{\textbf{Homo}} \\
\cmidrule(lr){2-5}\cmidrule(lr){6-14}
\textbf{Task}
& \textbf{Padding} & \textbf{FLoRA} & \textbf{Ours} & $\boldsymbol{\Delta}$
& \multicolumn{3}{c}{Rank = 4}
& \multicolumn{3}{c}{Rank = 8}
& \multicolumn{3}{c}{Rank = 16} \\
\cmidrule(lr){6-8}\cmidrule(lr){9-11}\cmidrule(lr){12-14}
& & & &
& \textbf{FedAvg} & \textbf{Ours} & $\boldsymbol{\Delta}$
& \textbf{FedAvg} & \textbf{Ours} & $\boldsymbol{\Delta}$
& \textbf{FedAvg} & \textbf{Ours} & $\boldsymbol{\Delta}$ \\
\midrule
\textsc{QNLI}
& 95.81 & 96.10 & \textbf{96.96} & +0.86
& 95.81 & \textbf{97.78} & +1.97
& 96.50 & \textbf{98.47} & +1.97
& 97.00 & \textbf{98.65} & +1.65 \\
\textsc{RTE}
& 95.45 & 97.20 & \textbf{99.71} & +2.51
& 85.94 & \textbf{88.89} & +2.95
& 85.62 & \textbf{90.05} & +4.43
& 86.10 & \textbf{90.80} & +4.70 \\
\textsc{QQP}
& 86.72 & 87.10 & \textbf{87.97} & +0.87
& 84.40 & \textbf{87.60} & +3.20
& 83.10 & \textbf{85.30} & +2.20
& 86.50 & \textbf{89.87} & +3.37 \\
\textsc{MNLI(matched)}
& 65.48 & 69.30 & \textbf{72.19} & +2.89
& 70.20 & \textbf{73.05} & +2.85
& 69.85 & \textbf{73.60} & +3.75
& 70.10 & \textbf{74.00} & +3.90 \\
\textsc{MNLI(mismatch)}
& 64.90 & 68.80 & \textbf{71.62} & +2.82
& 69.55 & \textbf{72.48} & +2.93
& 69.20 & \textbf{73.05} & +3.85
& 69.50 & \textbf{73.70} & +4.20 \\
\textsc{SST-2}
& 81.10 & 81.70 & \textbf{82.30} & +0.60
& 82.06 & \textbf{82.14} & +0.08
& \textbf{81.90} & 81.88 & $-0.02$
& 82.00 & \textbf{82.20} & +0.20 \\
\textsc{MMLU}
& 41.55 & 44.10 & \textbf{45.60} & +1.50
& 44.00 & \textbf{46.30} & +2.30
& 44.80 & \textbf{47.00} & +2.20
& 45.25 & \textbf{47.60} & +2.35 \\
\midrule
\textbf{Avg.}
& 75.86 & 77.76 & \textbf{79.48} & +1.72
& 75.99 & \textbf{78.32} & +2.33
& 75.85 & \textbf{78.48} & +2.63
& 76.64 & \textbf{79.55} & +2.91 \\
\bottomrule
\end{tabular}
\caption{Federated aggregation results under rank-heterogeneous and rank-homogeneous clients.
$\Delta$ denotes absolute improvement (percentage points) of \textbf{Ours} over the strongest baseline in each setting:
$\max(\text{FLoRA}, \text{Padding})$ for Hetero, and FedAvg for Homo (Rank=4/8/16).}
\label{tab:stacking_vs_baselines}
\end{table*}

\paragraph{Tasks and datasets.}
We evaluate federated parameter-efficient fine-tuning on GLUE-style and MMLU benchmarks~\cite{wang2018glue,hendrycksmeasuring}.
We report results on \textsc{QNLI}, \textsc{RTE}, \textsc{QQP}, \textsc{MNLI}, and \textsc{SST-2},
which cover natural language inference, paraphrase identification, and sentiment classification.
We use the official train/dev splits and report dev-set performance.

\paragraph{Backbone and adapter.}
We adopt LLaMA-7B~\cite{touvron2023llama} as the frozen backbone and insert LoRA adapters~\cite{hu2022lora} into linear projection layers.
Only LoRA parameters are optimized while the backbone weights remain fixed.

\paragraph{Federated protocol and rank heterogeneity.}
We simulate a cross-device FL setting with $K=8$ clients and run $T=30$ communication rounds.
Unless otherwise stated, all clients participate in each round and the server aggregates client updates with weights
proportional to local data sizes, i.e., $\alpha_k = n_k / \sum_j n_j$.
To model structural heterogeneity induced by rank mismatch, we set client ranks as
$\{r_k\}=[4,4,8,8,8,8,16,16]$.
When a homogeneous setting is required, we use a shared rank $r_k=r$ for all clients. In an additional scalability study, we vary the number of participating clients $K$ while keeping the remaining hyperparameters consistent.

\paragraph{Local optimization.}
Each client uses Adam with a fixed learning rate of $2\times 10^{-4}$ and batch size $128$.
Within each communication round, clients perform a fixed number of local update steps ($10$) on mini-batches before uploading their LoRA updates.
All remaining optimizer hyperparameters follow the default settings in our implementation.
These hyperparameters are kept identical across methods unless the experiment explicitly controls a specific factor, ensuring fair comparisons.%

\paragraph{Evaluation metrics.}
We report Accuracy on \textsc{QNLI}/\textsc{RTE}/\textsc{MNLI}/\textsc{SST-2}.
And for \textsc{QQP}, we also report Accuracy to maintain a unified metric across tasks.

\paragraph{Baselines.}
We compare the proposed aggregation strategy against representative alternatives under identical training budgets:
\textbf{(i) Zero-padding} aligns heterogeneous LoRA updates by padding lower-rank factors to the maximum rank before
weighted averaging;
\textbf{(ii) FedAvg} performs standard weighted averaging of LoRA parameters in the homogeneous-rank setting;
\textbf{(iii) Stacking(FLoRA)} concatenates client low-rank updates along the rank dimension to preserve the union of client subspaces.

\paragraph{Reproducibility.}
All experiments are conducted on a machine equipped with an Intel Xeon Gold 5318Y CPU, 128GB RAM,
and 2$\times$NVIDIA RTX A6000 GPUs, running Ubuntu 22.04.5, Python 3.10, and CUDA 12.4.
Key hyperparameters and system configurations are summarized in Table~\ref{tab:exp_config}.

\subsection{Results under Rank and Data Heterogeneity}
\label{sec:exp_main}

\paragraph{Rank heterogeneous clients.}
Table~\ref{tab:stacking_vs_baselines} reports results under mismatched client ranks.
Overall, \textbf{Ours} consistently outperforms zero-padding and FLoRA across all tasks.
For example, it improves \textsc{QNLI} from 95.81\% to 96.96\% and \textsc{QQP} from 86.72\% to 87.97\%, and achieves notably larger gains on \textsc{RTE} and \textsc{MNLI}, which are more sensitive to cross-client directional mismatch.

These improvements stem from how different methods handle structural mismatch under rank heterogeneity.
Zero-padding dilutes informative low-rank updates by enforcing a common high-rank parameterization, while FLoRA aggregates adapters in a monolithic manner.
By contrast, Ours performs selective aggregation only on the shared component, leading to more stable and reliable updates under mismatched ranks.

\paragraph{Homogeneous rank sanity check.}
To verify that the benefit of our method is not merely due to resolving dimensional incompatibility, we further compare it with FedAvg-style averaging when all clients use the same rank (Table~\ref{tab:stacking_vs_baselines}).
Even in this homogeneous setting, stacking remains consistently superior or comparable.

For rank $r=4$, Ours improves \textsc{QQP} by 3.20\%, \textsc{QNLI} by 1.97\%, and \textsc{RTE} by 2.95\%, while maintaining comparable performance on \textsc{SST-2}.
For rank $r=8$, Ours again improves \textsc{QQP} (2.20\%), \textsc{QNLI} (1.97\%), and \textsc{RTE} (4.43\%), with only a negligible difference on \textsc{SST-2} (-0.02\%).
For rank $r=16$, Ours yields further gains on \textsc{QQP} (3.37\%), \textsc{QNLI} (1.65\%), and \textsc{RTE} (4.70\%), while slightly improving performance on \textsc{SST-2} (0.20\%).
These results indicate that Ours acts as a subspace-preserving aggregator that remains beneficial even when the rank is fixed, rather than being a mere workaround for heterogeneous dimensions.

\begin{figure}[t]
  \centering
  \includegraphics[width=\linewidth]{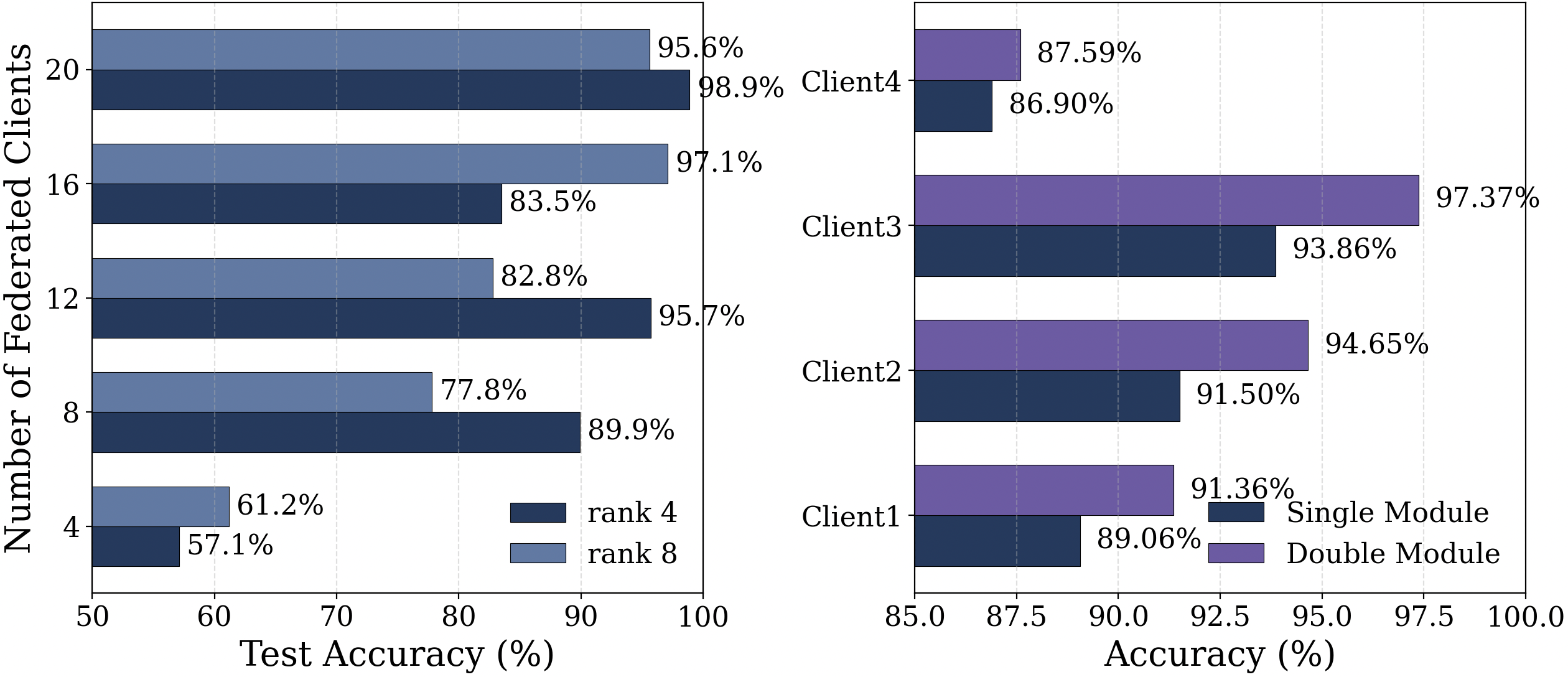}
  \caption{(Left) Test accuracy under varying numbers of federated clients ($K$) for two rank settings. Increasing $K$ generally improves performance due to broader data coverage, while non-IID update discrepancy can cause transient fluctuations.  (Right) Per-client accuracy comparison between single-module and dual-module adapters.}
  \label{fig:num_clients}
\end{figure}

\paragraph{Effect of client population size.}
We study the impact of the federated population size by varying the number of participating clients $K$ while keeping all other settings fixed.
As shown in Figure~\ref{fig:num_clients}, increasing $K$ generally improves test accuracy, reflecting the benefit of broader data coverage in larger federations.
With very few clients ($K{=}4$), aggregation is less effective due to limited diversity of update directions.

The performance trend is not strictly monotonic, and a temporary drop is observed at intermediate $K$ values (e.g., around $K{=}16$), which is consistent with non-IID federated optimization.
Comparing the two rank settings, lower ranks exhibit better robustness in the small-to-medium regime, whereas higher ranks achieve a higher accuracy ceiling as the client population grows.
Overall, these results indicate that our method scales well with the federation size while exhibiting a non-trivial interaction between rank and population size.

\begin{figure}[t]
    \centering
    \includegraphics[width=\columnwidth]{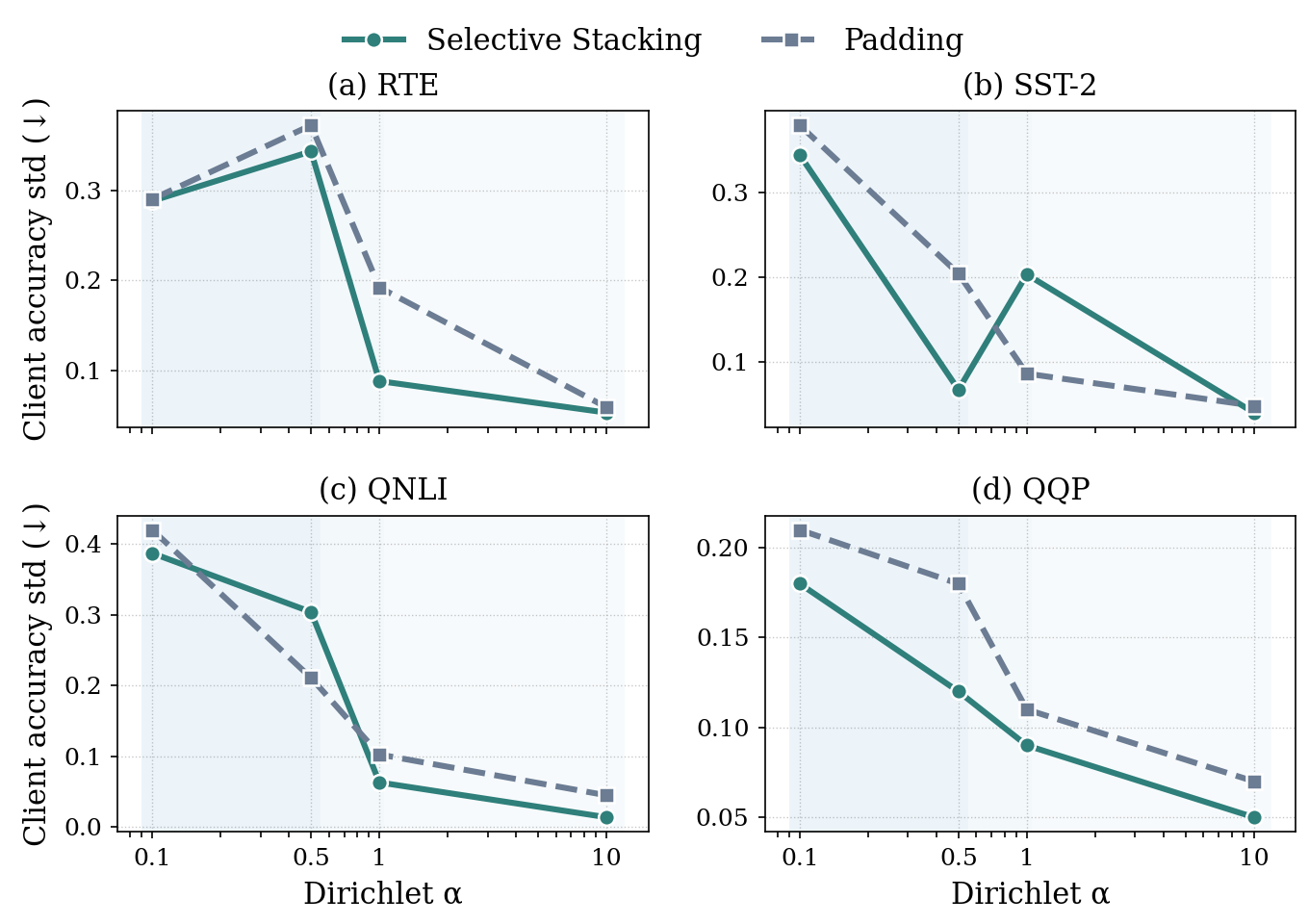}
    \caption{Client-level accuracy standard deviation under varying degrees of data heterogeneity (Dirichlet $\alpha$) for selective stacking vs. padding.}
    \label{fig:dual_vs_single}
\end{figure}

\paragraph{Robustness to data heterogeneity.}
We further examine robustness under Non-IID data partitions controlled by a Dirichlet parameter $\alpha$ using the client-level accuracy standard deviation in Figure~\ref{fig:dual_vs_single}.
Lower variance indicates more consistent performance across heterogeneous clients.

Under strong heterogeneity ($\alpha=0.1$), stacking achieves lower variance than padding on all tasks, e.g., on \textsc{SST-2} (0.344 vs.\ 0.379) and \textsc{QNLI} (0.387 vs.\ 0.420), indicating improved stability when client data distributions are highly skewed.
At moderate heterogeneity ($\alpha=0.5$), stacking again consistently reduces variance, for example lowering RTE from 0.372 to 0.343 and QNLI from 0.111 to 0.104, showing better robustness across clients.

At $\alpha=1.0$, stacking remains slightly more stable on \textsc{RTE} (0.088 vs.\ 0.092), but exhibits higher variance on \textsc{SST-2} and \textsc{QNLI}.
This suggests that when data heterogeneity is moderate, preserving a broader set of client-specific low-rank directions may amplify inter-client discrepancies on certain tasks, even though the shared model remains competitive.
Importantly, when the partition approaches near-IID ($\alpha=10$), stacking again achieves uniformly lower variance across all tasks, showing that stacking does not inherently harm stability.

\subsection{Privacy and Rank Sensitivity}
\label{sec:exp_privacy_rank}
\begin{figure}[t]
    \centering
    \includegraphics[width=\columnwidth]{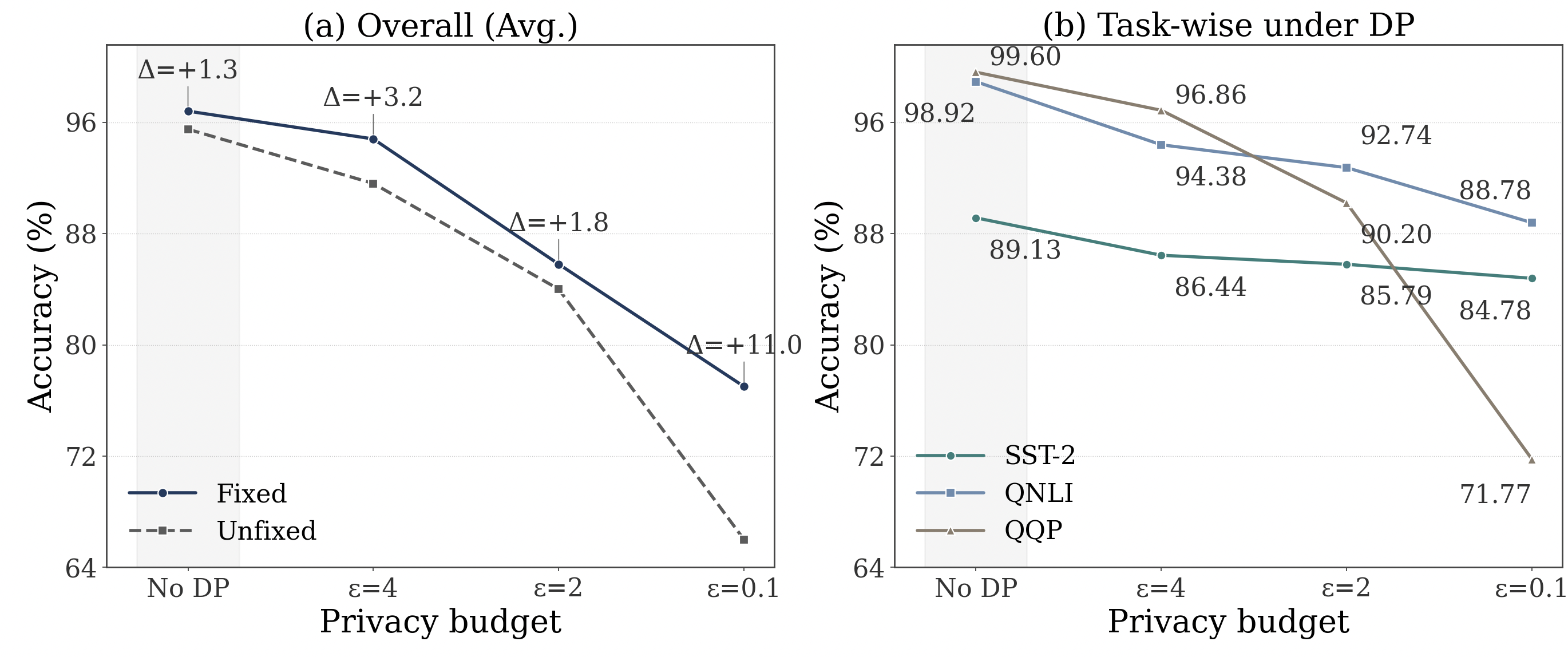}
    \caption{Accuracy under different privacy budgets $\epsilon$.
    Fixed denotes applying structural constraints (selective stacking and rank control) on the shared module,
    while Unfixed corresponds to unconstrained aggregation over all updates.
    The fixed design exhibits substantially better stability under strong privacy constraints.}
    \label{fig:dp_fixed}
\end{figure}

\paragraph{Privacy and utility trade-off.}
We apply DP-SGD during private training and report task accuracies under different privacy budgets $\epsilon$
(Figure~\ref{fig:dp_fixed}). As $\epsilon$ increases (weaker privacy), the required noise scale decreases and performance improves.

Figure~\ref{fig:dp_fixed} further shows that enforcing structural constraints on the shared module
significantly improves robustness under strong privacy constraints.
When $\epsilon$ is small, unconstrained aggregation suffers from severe performance degradation,
whereas the fixed shared-module design maintains substantially higher accuracy.

\paragraph{Rank sensitivity with and without DP.}
Table~\ref{tab:rank_sensitivity} analyzes how the DP noise affects performance.
Without DP, smaller ranks can be more stable on some tasks, while overly large ranks may introduce redundancy and hurt aggregation.
With DP ($\epsilon=1$), the optimal rank can shift due to noise, but small-to-medium ranks often remain a good trade-off.

\begin{table}[t]
\centering
\small
\setlength{\tabcolsep}{4pt}
\renewcommand{\arraystretch}{1.05}
\begin{tabular}{c l c c c c}
\toprule
Rank $r$ & Setting & \textsc{QNLI} & \textsc{MNLI (m)} & \textsc{MNLI (mm)} & \textsc{QQP} \\
\midrule
4  & No DP             & \textbf{94.39} & 74.62 & \textbf{89.63} & 86.02 \\
   & DP ($\epsilon=1$)  & 92.41          & 72.45 & 85.42          & 84.83 \\
\midrule
8  & No DP             & 88.78          & \textbf{79.60} & 79.22 & 82.06 \\
   & DP ($\epsilon=1$)  & 84.49          & 73.34          & 74.18 & 81.25 \\
\midrule
16 & No DP             & 84.06          & 71.04 & 76.93 & \textbf{89.87} \\
   & DP ($\epsilon=1$)  & 81.41          & 70.75 & 69.32 & 87.32 \\
\midrule
32 & No DP             & 88.89          & 66.46 & 73.41 & 80.55 \\
   & DP ($\epsilon=1$)  & 85.47          & 62.04 & 71.02 & 77.99 \\
\bottomrule
\end{tabular}
\caption{Rank sensitivity with and without differential privacy (DP).}
\label{tab:rank_sensitivity}
\end{table}

\subsection{Evidence for Decoupled Benefit}
\label{sec:exp_dual_evidence}
We further provide direct evidence that the dual-adapter (decoupled) design itself contributes to performance, independent of selective stacking and rank budgeting. As shown in Figure~\ref{fig:num_clients}
(Right), under matched training budgets the dual-adapter improves accuracy on every client, raising the average from 90.33\% to 92.74\% with an improvement of 2.41 percentage points. Per-client gains range from 0.69\% on Client~4 to 3.51\% on Client~3, indicating that the shared and private branches capture complementary low-rank subspaces rather than redundant capacity.

\section{Conclusion}
\label{sec:conclusion}

We studied federated fine-tuning of LLMs with LoRA under practical client heterogeneity, including differences in data distributions, system constraints, adapter ranks, and privacy requirements. While stacking-based aggregation provides a natural way to reconcile rank mismatch, we argue that stacking alone is insufficient, as it implicitly forces shared alignment on all client updates, suppressing personalization and becoming fragile under differential privacy noise.

We proposed the \textbf{SDFLoRA} framework, which structurally decomposes each client LoRA adapter into shared and private components. Building on this decomposition, we introduce selective stacking that applies stacking-based alignment only to the shared module while keeping private modules private and unaggregated. This configuration supports privacy-aware optimization by injecting DP noise exclusively into the shared module. Experiments on multiple benchmarks show that our approach outperforms representative federated LoRA baselines and achieves a better balance between utility and privacy under DP noise. In future work, we plan to extend this structural separation to other adapter families and investigate adaptive mechanisms for allocating shared versus private capacity based on client characteristics and task demands.

\section*{Acknowledgments}

This work was supported by the Key R\&D Program of Zhejiang Province under Grant No.~2025C01084 and the CAAI--MindSpore Open Fund, developed on the OpenI Community, under Contract No.~CAAIXSJLJJ 2025 MindSpore 04.

\bibliographystyle{named}
\bibliography{ijcai26}

\end{document}